%% file: Template.tex
\documentclass{article}
\usepackage{spconf,amsmath,graphicx}
\usepackage{multirow}
\usepackage{booktabs}
\usepackage[sort&compress,square,comma, numbers]{natbib}
\usepackage{color}
\usepackage{colortbl}
\usepackage{tcolorbox}
\usepackage{xcolor}
\newtcbox{\blue}[1][]{on line,boxsep=0.5pt,left=0pt,right=0pt,top=0pt,bottom=0pt,colframe=white,colback=blue!25!white,#1}
\newtcbox{\red}[1][]{on line,boxsep=0.5pt,left=0pt,right=0pt,top=0pt,bottom=0pt,colframe=white,colback=red!25!white,#1}

\title{MLPs Compass: What is learned when MLPs are combined with PLMs?}
\newcommand{\uestc}{$^1$}
\newcommand{\hust}{$^2$}

\name{Li Zhou\uestc\sthanks{We are grateful to Daniel Hershcovich for initial discussions about the research questions. Li Zhou is financially supported by the Fund of High Level Academic Conference Program in the Graduate School of UESTC.}, Wenyu Chen\uestc*\thanks{*Corresponding author: Wenyu Chen,
email: cwy@uestc.edu.cn}, Yong Cao\hust, Dingyi Zeng\uestc,  Wanlong Liu\uestc, Hong Qu\uestc} 

\address{{\uestc}University of Electronic Science and Technology of China\\
{\hust}Huazhong University of Science and Technology \\} 

\begin{document}
%
\maketitle
\begin{abstract}
While Transformer-based pre-trained language models and their variants exhibit strong semantic representation capabilities, the question of comprehending the information gain derived from the additional components of PLMs remains an open question in this field.
Motivated by recent efforts that prove Multilayer-Perceptrons (MLPs) modules achieving robust structural capture capabilities, even outperforming Graph Neural Networks (GNNs), 
this paper aims to quantify whether simple MLPs can further enhance the already potent ability of PLMs to capture linguistic information.
Specifically, we design a simple yet effective probing framework containing MLPs components based on BERT structure and conduct extensive experiments encompassing 10 probing tasks spanning three distinct linguistic levels. 
The experimental results demonstrate that MLPs can indeed enhance the comprehension of linguistic structure by PLMs.
Our research provides interpretable and valuable insights into crafting variations of PLMs utilizing MLPs for tasks that emphasize diverse linguistic structures.

\end{abstract}
\begin{keywords}
Pre-trained language models, Linguistic structures, Multilayer Perceptron, Interpretation, Probing
\end{keywords}
\section{Introduction}
\label{sec:intro}
The landscape of natural language processing (NLP) has been revolutionized by large pre-trained language models (PLMs) based on transformer architecture, significantly advancing the state of the art in numerous NLP domains.
To comprehend the workings of PLMs, some research endeavors~\cite{,vulic-etal-2020-probing, NEURIPS2021_86b3e165, wang-etal-2020-chinese-bert, 10.1162/tacl_a_00349} focus on conducting interpretable explorations.
Some works~\cite{jawahar-etal-2019-bert, niu-etal-2022-bert} delve into  BERT's implicit understanding of linguistic structure through its representations, revealing its ability to capture a diverse hierarchy of linguistic information. Furthermore, researchers~\cite{tenney-etal-2019-bert} demonstrate that the pre-trained language model BERT elucidates the constituents of the conventional NLP pipeline in an interpretable and localizable manner. They furnish fresh evidence affirming that deep language models can embody the sorts of syntactic and semantic abstractions traditionally considered essential for language processing. The above works provides an explanation for BERT's remarkable performance across a wide range of tasks.

Except for semantic representation, many studies~\cite{zhou2020weighted, cao2023geo, cao-etal-2023-pay, zhou2023revisiting, liu2022document, liu2023enhancing} focus on intricate frameworks to integrate structural features for semantically relevant tasks,
such as relation extraction, by integrating the dependency structure of text with Graph Neural Networks (GNNs)~\cite{zeng2022simple}.
GNNs can capture both topological-structural and feature-related information for graph representation learning~\cite{ZHOU2023110377,zeng2023rethinking}.
Interestingly, current studies~\cite{galke-scherp-2022-bag} proved that using MLPs can effectively and efficiently capture structural features, even surpass GNNs in some tasks. To prove this, we conduct an experiment on two relation extraction tasks as illustrated in Table~\ref{tab:RE-task}, where we can observe that PLMs still demonstrate improved performance on both benchmark datasets, i.e. ReTACRED and SemEval, by fusing extra MLPs representations.

\begin{table}[]
\centering
\scalebox{0.8}{
\begin{tabular}{@{}lrr@{}}
\toprule
                  & \textbf{ReTACRED} & \textbf{SemEval} \\ \midrule
\textbf{BERT}     & 87.66±0.18        & 91.07±0.26       \\
\textbf{BERT+MLPs} & 88.05±0.21        & 91.31±0.23       \\ \bottomrule
\end{tabular}}
\caption{Performance impact of applying MLPs without structural bias on PLMs. ReTACRED \cite{stoica2021re} and SemEval\cite{hendrickx-etal-2010-semeval} are two popular datasets for relation extraction.}
\label{tab:RE-task}
\end{table}


MLPs are a foundational neural network component in model design, playing a crucial role in various adaptations. 
Several studies highlight that even basic MLPs possess the capability to uncover latent semantic information~\cite{li-etal-2023-well, anonymous2023rethinking}, exhibit greater transferability in unsupervised pretraining compared to supervised pretraining methods~\cite{wang2022revisiting}. However, what is learned when MLPs are combined with powerful PLMs is still an open question.
Therefore, we propose a simple yet effective probing framework containing extra MLP components based on BERT structure and introduce 10 probing tasks across 3 linguistic levels. Our objective is to elucidate the reasons behind the performance improvement brought about by MLPs that does not introduce structural bias. The specific research questions are as follows:
\begin{description}
\setlength{\itemsep}{0pt}
\setlength{\parsep}{0pt}
\setlength{\parskip}{0pt}
    \item[RQ1.] What can be learned when basic MLPs are integrated with the transformer structure in PLMs?
    \item[RQ2.] Does layer sensitivity exist in the performance changes when combining MLPs and PLM?
    \item[RQ3.] In the enhancement of PLMs with MLPs, which aspect of linguistic information understanding is MLPs particularly skilled at improving?
\end{description}
Our experiments demonstrate that when combined with MLPs components devoid of any structural bias, PLMs can indeed enhance language structure comprehension, 
encompassing surface, syntactic, and semantic levels. Our research offers interpretable and valuable insights into the utilization of MLPs in creating PLM variants tailored for tasks that emphasize distinct language structures.

\section{Preliminaries and Related Works}
\subsection{Transformer-based Structure}
\vspace{-2pt}
The Transformer-based structure~\cite{vaswani2017attention} 
serves as a general-purpose feature encoder for most PLMs.  Transformers are typically composed of multiple layers, each comprising a multi-head self-attention mechanism and a feedforward neural network, among other components. Stacking these layers empowers the model to acquire progressively intricate features and relationships. Specifically, for a given input sequence $X=\left[ x_0, x_1, …, x_{n-1} \right] $ that have been tokenized into subtoken units, the deep encoder of Transformer-based PLMs generates a series of representations from various layers: $\left[ \mathbf{H}^{\left( 0 \right)}, \mathbf{H}^{\left( 1 \right)}, …, \mathbf{H}^{\left( L-1 \right)} \right]$, where $\mathbf{H}^{\left( l \right)}=\left( \boldsymbol{h}_{0}^{\left( l \right)}, \boldsymbol{h}_{1}^{\left( l \right)}, …, \boldsymbol{h}_{n-1}^{\left( l \right)} \right) $ denotes the representation learned by the $l_{th}$ encoder layer.




\vspace{-5pt}
\subsection{Interpretability of PLMs}
\vspace{-2pt}
BERT~\cite{devlin-etal-2019-bert}, as a representative of PLMs, captures contextual word meaning bidirectionally through extensive pre-training on large text corpora. It proves highly effective in various NLP tasks, showcasing the versatility of the Transformer-based architecture in NLP. Some efforts are dedicated to elucidating the reasons behind BERT's remarkable capabilities. The attention in BERT has been demonstrated to reflect syntactic structures~\cite{ravishankar-etal-2021-attention}.
BERT representations are showed to be hierarchical rather than linear~\cite{liu-etal-2019-linguistic}, the embeddings of BERT can encode information about parts of speech, syntactic chunks and roles~\cite{tenney-etal-2019-bert}, and BERT contains important syntactic information~\cite{niu-etal-2022-using}. Except English language, Chinese BERT based on the same Transformer structure can also \cite{wang-etal-2020-chinese-bert} capture the word structure.

\vspace{-2pt}
\section{methodology}
We utilize a performance-based probing methodology, where an auxiliary task is employed to assess the existence of certain types of knowledge. This involves training a supervised classifier using solely BERT's representation as input, and achieving satisfactory classifier performance serves as evidence of the presence of pertinent linguistic knowledge.
To explain and quantify the information learned by the combination of MLPs with PLMs, we propose a probing framework and introduce three types of probing tasks.

\begin{figure}[h]
    \centering
    \includegraphics[width=0.85\linewidth]{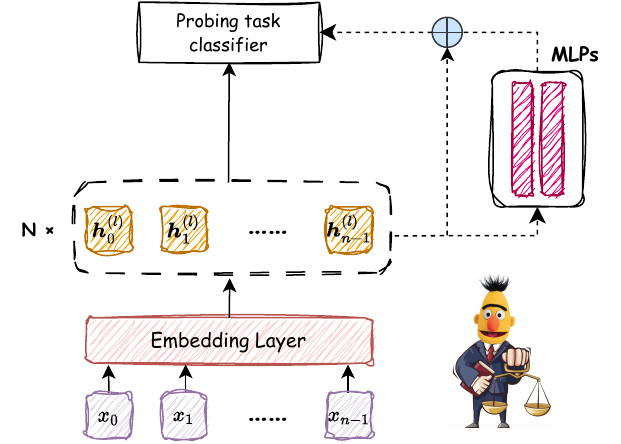}
    \caption{Probing framework. All parameters inside the dashed line and the embedding layer are fixed.}
    \label{fig:probing_framework}
\end{figure}

\vspace{-5pt}

\label{sec:format}
\subsection{Probing Framework}
\vspace{-3pt}
Our probing framework is illustrated in Figure~\ref{fig:probing_framework}. For each probe task, we train a probe classifier $P_{\tau}$. However, unlike the conventional usage of BERT, we keep all encoder weights frozen instead of fine-tuning BERT during training.
This ensures that the encoder doesn't adapt its internal representations specifically for the probe task. 
To study the acquired information within each encoder layer, we examine each layer individually, and the probing process, represented by the solid arrows on the left side of Figure~\ref{fig:probing_framework}, is detailed as follows:
\begin{equation}
\vspace{-3pt}
    \hat{Y}_{\tau}^{\left( l \right)}=P_{\tau}^{\left( l \right)}\left( \mathbf{H}^{\left( l \right)} \right)
\vspace{-3pt},
\end{equation}
where $\hat{Y}_{\tau}^{\left( l \right)}$ is the prediction for the probe task $\tau$ based on the representation of the $l_{th}$ encoder layer. 

To investigate the information learned through the combination of MLPs with PLMs, we introduce an MLPs block between the probe classifier and BERT representations. Concurrently, we utilize ResNet to integrate the initial features, ensuring that the model leverages both the original BERT representation and the MLP-acquired features. The MLPs probing process is illustrated in the dashed arrow section of Figure~\ref{fig:probing_framework} and can be represented by the following eequation.
\begin{equation}
\vspace{-3pt}
    \hat{Y}_{\tau}^{\left( l \right)}=P_{\tau}^{\left( l \right)}\left( \mathrm{MLPs}\left( \mathbf{H}^{\left( l \right)} \right) +\mathbf{H}^{\left( l \right)} \right)
\vspace{-2pt}
\end{equation}

By employing probing designs with or without MLPs, we can attribute the performance discrepancy between the two trained probing classifiers to the incorporation of MLPs.~\footnote{We conduct all our probing tasks at the sentence level, using $h^{(l)}_0$ as the input instead of $\mathbf{H}^{\left( l \right)}$.}

\input{table/overall_results}
\vspace{-5pt}
\subsection{Probing Tasks}
\vspace{-3pt}
We apply SentEval~\cite{conneau-etal-2018-cram, conneau-kiela-2018-senteval} for our probing tasks, encompassing 10 sentence-level probing tasks across three linguistic levels: surface, syntactic, and semantic.

\textbf{Surface tasks.}\quad Surface tasks assess the degree to which sentence embeddings retain the surface properties of the encoded sentences. Solving the surface tasks requires examining the tokens in the input sentences, without the need for in-depth linguistic knowledge. There are two surface tasks: predicting the length of a sentence based on the number of words (\textit{(SentLen)}) and detecting the possibility of recovering the original words in the sentence from its embeddings (\textit{WC}).

\textbf{Syntactic tasks.}\quad These tasks are designed to assess whether sentence embeddings exhibit sensitivity to the syntactic properties of the sentences they encode. Specifically, we probe for sensitivity to legal word order (\textit{BShift}), the depth of the syntactic tree (\textit{TreeDepth}), and the sequence of top-level constituents in the syntactic tree (\textit{TopConst}).

\textbf{Semantic tasks.}\quad Semantic tasks, in addition to relying on syntactic structure, demand an understanding of the meaning conveyed by a sentence.
The \textit{Tense} task involves identifying the tenses of the main-clause verb. The \textit{SubjNum} task and the \textit{ObjNum} task center on determining the number of subjects and the number of direct objects of the main clause, respectively. 
Furthermore, we also probe for the sensitivity to random noun/verb replacement (\textit{SOMO}) and the random swapping of coordinated clausal conjuncts (\textit{CoordInv}).

For each task, there are 100k sentences for training and 10k sentences each for validation and testing, respectively. It's worth noting that all sets are balanced, ensuring an equal number of instances for each target class.

\section{experiments}
\label{sec:pagestyle}

\begin{figure*}[ht]
    \centering
    \includegraphics[width=0.92\linewidth]{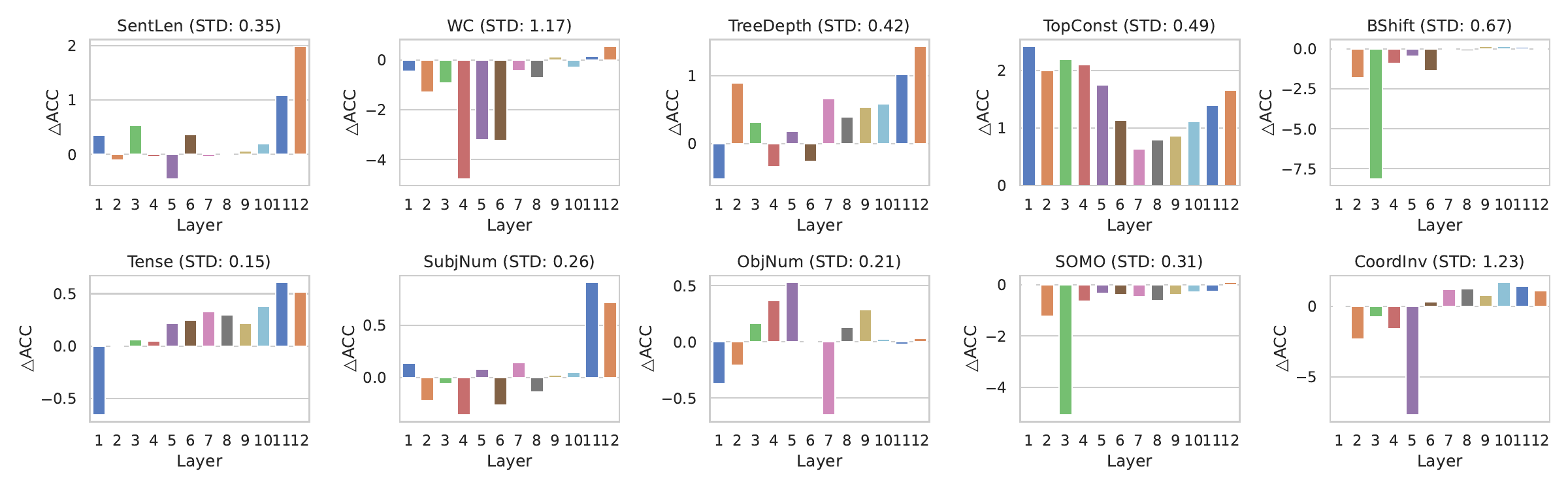}
    \caption{The performance variations across different layers for various probing tasks. A positive value indicates performance improvement, whereas a negative value indicates performance degradation. STD denotes the standard deviation of the performance differences across layers (excluding the maximum and minimum values).}
    \label{fig:layer}
\end{figure*}

\subsection{Experimental Setup}
\vspace{-3pt}
We utilize \texttt{BERT-base} with 12 layers as our foundational PLM for probing MLPs. 
Using our proposed probing framework, we conduct comparative experiments by including~\footnote{MLPs adopt two layers, aligning with the typical number of layers in most GNNs.} or excluding MLPs components in the probing process.
For each probing task, we conduct training using the Adam optimizer with a batch size of 64 for a total of 4 epochs. Additionally, we implement an early stopping mechanism based on the validation set, with a patience of 5.
We report \texttt{Accuracy (ACC)} on the test set to evaluate the amount of information learned. To ensure the reliability of our experimental results, we run each experiment with 5 different random seeds.

\vspace{-7pt}
\subsection{Layer-wise Results: RQ1}
\vspace{-3pt}
To study what has been enhanced learned by MLPs, we compare the probing results for different layers with and without the MLPs component.
As shown in Table~\ref{tab:results}, 
in most layers of the probing experiments, combining MLPs with PLM can improve the performance of the probing tasks at three different levels. This demonstrates that MLPs indeed enhance linguistic structure comprehension of PLMs, even without adding any structural bias.
Besides, we find that it is easier to show consistent improvements based on high-layer representations.
Although MLPs may introduce different changes in information at various layers, which can either enhance or diminish it, the fundamental hierarchical structure remains consistent before and after the introduction of MLPs: the lower layers focus on surface information, the middle to upper layers emphasize syntactic information and semantic information.





\vspace{-7pt}
\subsection{Layer Sensitivity: RQ2}
\vspace{-3pt}
To investigate the layer sensitivity in the performance changes 
when combining MLPs and PLM, we visualize the performance variations across layers for different probing tasks in Figure~\ref{fig:layer}, and show the standard deviation (STD) of performance differences across all layers.
We can find that the ability of MLPs to capture additional language information varies across BERT's middle and lower-level layers, while consistently proving beneficial in its higher layers. This is supported by the fluctuating bars in the middle and lower-level layers but consistently positive results in high layers across almost all tasks.
Besides,  we also can conclude that MLPs' sensitivity to different layers is relatively moderate, as indicated by the low STD observed across most tasks.

\vspace{-7pt}
\subsection{Linguistic Information Comparison: RQ3}
\vspace{-3pt}
To analyze which type of language information MLPs excel at, we further use the test data representations in the same task group to conduct k-means clustering under two settings: with and without MLPs components. We evaluate the resulting clusters with Normalized Mutual Information (NMI)\footnote{NMI takes values between 0 and 1, with 0 indicating no mutual information (no agreement between the ground truth and predicted clusters) and 1 indicating perfect agreement.}. Table~\ref{tab:aspect} shows that the presence of MLPs components enhances the clustering performance across all three task categories, indicating that even basic MLPs are capable of acquiring surface, syntactic, and semantic information.
In particular, MLPs are better at capturing both syntactic and semantic information, as evidenced by their more significant improvements in the cluster task compared to surface one. This observation helps elucidate the phenomenon illustrated in Table~\ref{tab:RE-task}, whereby the inclusion of straightforward MLPs leads to enhanced performance in relation extraction, even in the absence of structural bias.

\input{figure/aspec}


\vspace{-7pt}
\section{Conclusion}
\label{sec:con}
In this paper, we introduce a straightforward yet effective probing framework to investigate the information learned by MLPs in combination with PLM. Our extensive experiments, encompassing 10 probing tasks spanning 3 linguistic levels, demonstrate the superior performance of our proposed framework. 
Experimental results indicate that MLPs can boost PLMs in capturing additional surface, syntactic, and semantic information, with a stronger capacity for enhancing the latter two. 
Moreover, when leveraging high-layer representations from PLMs, MLPs exhibit a greater ability to acquire additional information. 
Our work provides interpretable and valuable insights into crafting variations of PLMs utilizing MLPs for tasks that emphasize diverse linguistic structures.



\ninept
\bibliographystyle{IEEEbib}
\bibliography{anthology, strings,refs}

\end{document}

%% file: table/overall_results.tex
\begin{table*}[ht]
\centering
\scalebox{0.63}{

\begin{tabular}{@{}l|rrrrrrrrrr@{}}

\toprule
\multirow{3}{*}{\textbf{Layers}} & \multicolumn{4}{c|}{\textbf{Surface}}                                                              & \multicolumn{6}{c}{\textbf{Syntactic}}                                                                                              \\ \cmidrule(l){2-11} 
                                 & \multicolumn{2}{c}{\textbf{SentLen (6)}} & \multicolumn{2}{c|}{\textbf{WC (1000)}}                 & \multicolumn{2}{c}{\textbf{TreeDepth (7)}} & \multicolumn{2}{c}{\textbf{TopConst (20)}} & \multicolumn{2}{c}{\textbf{BShift (2)}}   \\ \cmidrule(l){2-11} 
                                 & \textbf{w/o}    & \textbf{w}             & \textbf{w/o} & \multicolumn{1}{r|}{\textbf{w}}          & \textbf{w/o}     & \textbf{w}              & \textbf{w/o}     & \textbf{w}              & \textbf{w/o}     & \textbf{w}             \\ \midrule
\textbf{1}                       & 85.83±0.95      & \textbf{86.19±1.17}    & 0.56±0.05    & \multicolumn{1}{r|}{0.12±0.04}           & 31.60±0.58       & 31.09±1.17              & 46.12±0.28       & \textbf{48.54±0.16}     & 50.00±0.00       & 50.01±0.01             \\
\textbf{2}                       & \blue{91.60±0.40}      & \blue{91.49±1.35}             & 2.35±0.10    & \multicolumn{1}{r|}{1.06±0.08}           & 34.68±0.59       & \blue{\textbf{35.58±0.29}}     & 58.19±0.41       & \textbf{60.2±0.43}      & 51.81±1.05       & 50.00±0.00             \\
\textbf{3}                       & \red{92.31±0.48}      & \red{\textbf{92.85±0.56}}    & 1.50±0.17    & \multicolumn{1}{r|}{0.58±0.05}           & 33.98±0.37       & \textbf{34.3±0.38}      & 56.77±0.18       & \textbf{58.97±0.65}     & 58.13±1.78       & 50.00±0.00             \\
\textbf{4}                       & 89.70±0.79      & 89.66±0.58             & \red{19.83±0.71}   & \multicolumn{1}{r|}{\blue{15.05±0.83}}          & 33.08±0.45       & 32.74±1.60              & 54.50±0.40       & \textbf{56.60±0.51}     & 69.74±1.47       & 68.83±2.12             \\
\textbf{5}                       & 85.00±0.72      & 84.55±0.78             & \blue{19.47±0.62}   & \multicolumn{1}{r|}{\red{16.26±0.81}}          & 33.90±0.97       & \textbf{34.08±0.76}     & 73.93±0.11       & \textbf{75.69±0.49}     & 78.44±0.32       & 77.99±0.40             \\
\textbf{6}                       & 81.10±0.81      & \textbf{81.46±0.49}    & 13.79±0.47   & \multicolumn{1}{r|}{10.57±0.74}          & \red{35.22±0.38}       & 34.97±1.36              & 78.86±0.13       & \textbf{80.0±0.50}      & 80.68±0.14       & 79.33±1.11             \\
\textbf{7}                       & 78.52±0.86      & 78.47±0.66             & 10.33±0.30   & \multicolumn{1}{r|}{9.90±0.33}           & \blue{34.98±0.53}       & \red{\textbf{35.64±0.56}}     & \red{80.32±0.15}       & \red{\textbf{80.96±0.10}}     & 81.25±0.14       & \textbf{81.33±0.17}    \\
\textbf{8}                       & 76.99±1.06      & \textbf{77.01±1.17}    & 7.99±0.15    & \multicolumn{1}{r|}{7.27±0.19}           & 34.15±0.44       & \textbf{34.54±0.22}     & \blue{79.55±0.20}       & \textbf{80.35±0.34}     & 81.98±0.25       & 81.86±0.29             \\
\textbf{9}                       & 74.15±0.45      & \textbf{74.21±0.96}    & 9.14±0.08    & \multicolumn{1}{r|}{\textbf{9.27±0.20}}  & 34.06±0.36       & \textbf{34.60±0.34}     & 79.52±0.24       & \blue{\textbf{80.38±0.32}}     & 85.51±0.19       & \textbf{85.70±0.13}    \\
\textbf{10}                      & 72.82±0.21      & \textbf{73.01±0.88}    & 9.41±0.16    & \multicolumn{1}{r|}{9.11±0.36}           & 33.72±0.66       & \textbf{34.31±0.33}     & 78.76±0.23       & \textbf{79.87±0.26}     & 85.72±0.18       & \textbf{85.90±0.09}    \\
\textbf{11}                      & 68.88±0.32      & \textbf{69.96±0.89}    & 10.59±0.28   & \multicolumn{1}{r|}{\textbf{10.75±0.28}} & 32.75±0.32       & \textbf{33.76±0.77}     & 77.02±0.15       & \textbf{78.42±0.28}     & \blue{85.86±0.15}       & \blue{\textbf{85.98±0.19}}    \\
\textbf{12}                      & 64.35±0.26      & \textbf{66.34±0.89}    & 14.26±0.24   & \multicolumn{1}{r|}{\textbf{14.82±0.54}} & 31.39±0.39       & \textbf{32.82±0.46}     & 72.86±0.16       & \textbf{74.52±0.13}     & \red{86.13±0.08}       & \red{\textbf{86.20±0.30}}    \\ \midrule \midrule
\multirow{3}{*}{\textbf{Layers}} & \multicolumn{10}{c}{\textbf{Semantic}}                                                                                                                                                                                                   \\ \cmidrule(l){2-11} 
                                 & \multicolumn{2}{c}{\textbf{Tense (2)}}   & \multicolumn{2}{c}{\textbf{SubjNum (2)}}                & \multicolumn{2}{c}{\textbf{ObjNum (2)}}    & \multicolumn{2}{c}{\textbf{SOMO (2)}}      & \multicolumn{2}{c}{\textbf{CoordInv (2)}} \\ \cmidrule(l){2-11} 
                                 & \textbf{w/o}    & \textbf{w}             & \textbf{w/o} & \textbf{w}                               & \textbf{w/o}     & \textbf{w}              & \textbf{w/o}     & \textbf{w}              & \textbf{w/o}     & \textbf{w}             \\ \midrule
\textbf{1}                       & 78.58±0.25      & 77.92±0.47    & 73.39±0.41   & \textbf{73.53±0.18}                      & 71.08±0.46       & 70.70±0.75               & 49.98±0.13       & 49.97±0.13              & 50.00±0.00       & 50.00±0.00             \\
\textbf{2}                       & 84.34±0.27      & 84.33±0.54             & 79.02±0.20   & 78.80±0.23                               & 77.31±0.67       & 77.11±1.18              & 51.20±1.08       & 49.97±0.13              & 52.31±1.21       & 50.00±0.00             \\
\textbf{3}                       & 85.45±0.30      & \textbf{85.51±0.37}    & 79.44±0.13   & 79.38±0.20                               & 76.27±1.43       & \textbf{76.44±0.76}     & 55.04±0.49       & 49.97±0.13              & 50.74±0.95       & 50.00±0.00             \\
\textbf{4}                       & 86.33±0.34      & \textbf{86.37±0.49}    & 79.51±0.23   & 79.15±0.47                               & 77.73±0.90       & \textbf{78.10±0.09}     & 57.88±0.14       & 57.23±0.35              & 51.59±0.94       & 50.00±0.00             \\
\textbf{5}                       & 88.63±0.16      & \textbf{88.85±0.29}    & 83.40±0.43   & \textbf{83.48±0.40}                      & 78.48±0.60       & \textbf{79.01±0.27}     & 59.33±0.30       & 58.98±0.48              & 57.72±1.15       & 50.01±0.01             \\
\textbf{6}                       & 88.60±0.28      & \textbf{88.85±0.27}    & \blue{86.34±0.24}   & \blue{86.08±0.91}                               & 79.12±0.62       & \textbf{79.13±0.50}     & 59.68±0.12       & 59.29±0.28              & 63.73±1.14       & \textbf{64.07±0.51}    \\
\textbf{7}                       & 88.86±0.18      & \textbf{89.19±0.25}    & 85.76±0.29   & \textbf{85.91±0.47}                      & \red{79.73±0.48}       & 79.08±0.19              & 60.42±0.37       & 59.94±0.49              & 69.66±1.05       & \textbf{70.86±0.95}    \\
\textbf{8}                       & \blue{89.16±0.14}      & \blue{\textbf{89.46±0.29}}    & 85.96±0.32   & 85.82±0.60                               & 79.02±0.26       & \blue{\textbf{79.15±0.33}}     & 60.32±0.42       & 59.68±0.61              & 71.14±0.86       & \textbf{72.41±0.57}    \\
\textbf{9}                       & \red{89.21±0.08}      & \textbf{89.43±0.26}    & \red{86.66±0.11}   & \red{\textbf{86.69±0.23}}                      & \blue{79.21±0.40}       & \red{\textbf{79.50±0.12}}     & 62.37±0.14       & 61.96±0.31              & \blue{73.74±0.82}       & \textbf{74.53±0.77}    \\
\textbf{10}                      & 89.10±0.08      & \red{\textbf{89.47±0.21}}    & 85.98±0.26   & \textbf{86.03±0.14}                      & 78.14±0.26       & \textbf{78.17±0.38}     & 62.70±0.19       & 62.41±0.34              & \red{73.82±1.17}       & \red{\textbf{75.52±0.86}}    \\
\textbf{11}                      & 88.86±0.31      & \blue{\textbf{89.46±0.20}}    & 83.56±0.50   & \textbf{84.47±0.25}                      & 77.09±0.23       & 77.07±0.41              & \blue{63.55±0.15}       & \blue{63.28±0.30}              & 73.27±0.53       & \blue{\textbf{74.68±0.65}}    \\
\textbf{12}                      & 88.87±0.27      & \textbf{89.39±0.11}    & 82.26±0.18   & \textbf{82.97±0.44}                      & 77.88±0.22       & \textbf{77.91±0.31}     & \red{64.00±0.21}       & \red{\textbf{64.09±0.20}}     & 71.25±0.69       & \textbf{72.38±0.52}    \\ \bottomrule
\end{tabular}}
\caption{The probing results from different layers of BERT-base. ``w/o'' and ``w'' respectively denote the absence and presence of MLPs component in our framework. Bold indicates an \textbf{improvement} in performance when MLP is combined with BERT. Red marks \red{the top-performing layer}, and blue denotes \blue{the second best} across different settings for various probing tasks.}
\label{tab:results}
\end{table*}

%% file: figure/aspec.tex
\begin{table}[]
\centering
\scalebox{0.85}{
\begin{tabular}{@{}l|rrr@{}}
\toprule
                   & \textbf{Surface} & \textbf{Syntactic} & \textbf{Semantic} \\ \midrule
\textbf{NMI (w/o)} & 0.60             & 0.14               & 0.07              \\
\textbf{NMI (w)}   & 0.66             & 0.57               & 0.49              \\ \midrule
\textbf{$\Delta$NMI}      & 0.06 (↑)             & 0.43 (↑)               & 0.42 (↑)              \\ \bottomrule
\end{tabular}}
\caption{Clustering performance with Normalized Mutual Information (NMI).}
\label{tab:aspect}
\vspace{-2mm}
\end{table}

%% file: Template.bbl
\begin{thebibliography}{10}

\bibitem{vulic-etal-2020-probing}
Ivan Vuli{\'c}, Edoardo~Maria Ponti, Robert Litschko, Goran Glava{\v{s}}, and
  Anna Korhonen,
\newblock ``Probing pretrained language models for lexical semantics,''
\newblock in {\em Proc. of EMNLP}, 2020.

\bibitem{NEURIPS2021_86b3e165}
Colin Wei, Sang~Michael Xie, and Tengyu Ma,
\newblock ``Why do pretrained language models help in downstream tasks? an
  analysis of head and prompt tuning,''
\newblock in {\em Proc. of NeurIPS}, 2021.

\bibitem{wang-etal-2020-chinese-bert}
Yile Wang, Leyang Cui, and Yue Zhang,
\newblock ``Does {C}hinese {BERT} encode word structure?,''
\newblock in {\em Proc. of COLING}, 2020.

\bibitem{10.1162/tacl_a_00349}
Anna Rogers, Olga Kovaleva, and Anna Rumshisky,
\newblock ``{A Primer in BERTology: What We Know About How BERT Works},''
\newblock {\em Transactions of the Association for Computational Linguistics},
  2021.

\bibitem{jawahar-etal-2019-bert}
Ganesh Jawahar, Beno{\^\i}t Sagot, and Djam{\'e} Seddah,
\newblock ``What does {BERT} learn about the structure of language?,''
\newblock in {\em Proc. of ACL}, 2019.

\bibitem{niu-etal-2022-bert}
Jingcheng Niu, Wenjie Lu, and Gerald Penn,
\newblock ``Does {BERT} rediscover a classical {NLP} pipeline?,''
\newblock in {\em Proc. of COLING}, 2022.

\bibitem{tenney-etal-2019-bert}
Ian Tenney, Dipanjan Das, and Ellie Pavlick,
\newblock ``{BERT} rediscovers the classical {NLP} pipeline,''
\newblock in {\em Proc. of ACL}, 2019.

\bibitem{zhou2020weighted}
Li~Zhou, Tingyu Wang, Hong Qu, Li~Huang, and Yuguo Liu,
\newblock ``A weighted gcn with logical adjacency matrix for relation
  extraction,''
\newblock in {\em ECAI 2020}. 2020.

\bibitem{cao2023geo}
Yong Cao, Ruixue Ding, Boli Chen, Xianzhi Li, Min Chen, Daniel Hershcovich,
  Pengjun Xie, and Fei Huang,
\newblock ``Geo-encoder: A chunk-argument bi-encoder framework for chinese
  geographic re-ranking,''
\newblock {\em arXiv:2309.01606}, 2023.

\bibitem{cao-etal-2023-pay}
Yong Cao, Xianzhi Li, Huiwen Liu, Wen Dai, Shuai Chen, Bin Wang, Min Chen, and
  Daniel Hershcovich,
\newblock ``Pay more attention to relation exploration for knowledge base
  question answering,''
\newblock in {\em Proc. of ACL Findings}, 2023.

\bibitem{zhou2023revisiting}
Li~Zhou, Wenyu Chen, Dingyi Zeng, Hong Qu, and Daniel Hershcovich,
\newblock ``Revisiting graph meaning representations through decoupling
  contextual representation learning and structural information propagation,''
\newblock {\em arXiv preprint arXiv:2310.09772}, 2023.

\bibitem{liu2022document}
Wanlong Liu, Li~Zhou, Dingyi Zeng, and Hong Qu,
\newblock ``Document-level relation extraction with structure enhanced
  transformer encoder,''
\newblock in {\em Proc. of IJCNN}, 2022.

\bibitem{liu2023enhancing}
Wanlong Liu, Shaohuan Cheng, Dingyi Zeng, and Hong Qu,
\newblock ``Enhancing document-level event argument extraction with contextual
  clues and role relevance,''
\newblock {\em arXiv preprint arXiv:2310.05991}, 2023.

\bibitem{zeng2022simple}
Dingyi Zeng, Li~Zhou, Wanlong Liu, Hong Qu, and Wenyu Chen,
\newblock ``A simple graph neural network via layer sniffer,''
\newblock in {\em Proc. of ICASSP}, 2022.

\bibitem{ZHOU2023110377}
Li~Zhou, Wenyu Chen, Dingyi Zeng, Shaohuan Cheng, Wanlong Liu, Malu Zhang, and
  Hong Qu,
\newblock ``Dpgnn: Dual-perception graph neural network for representation
  learning,''
\newblock {\em Knowledge-Based Systems}, 2023.

\bibitem{zeng2023rethinking}
Dingyi Zeng, Wenyu Chen, Wanlong Liu, Li~Zhou, and Hong Qu,
\newblock ``Rethinking random walk in graph representation learning,''
\newblock in {\em Proc. of ICASSP}, 2023.

\bibitem{galke-scherp-2022-bag}
Lukas Galke and Ansgar Scherp,
\newblock ``Bag-of-words vs. graph vs. sequence in text classification:
  Questioning the necessity of text-graphs and the surprising strength of a
  wide {MLP},''
\newblock in {\em Proc. of ACL}, 2022.

\bibitem{stoica2021re}
George Stoica, Emmanouil~Antonios Platanios, and Barnab{\'a}s P{\'o}czos,
\newblock ``Re-tacred: Addressing shortcomings of the tacred dataset,''
\newblock in {\em Proc. of AAAI}, 2021.

\bibitem{hendrickx-etal-2010-semeval}
Iris Hendrickx, Su~Nam Kim, Zornitsa Kozareva, Preslav Nakov, Diarmuid
  {\'O}~S{\'e}aghdha, Sebastian Pad{\'o}, Marco Pennacchiotti, Lorenza Romano,
  and Stan Szpakowicz,
\newblock ``{S}em{E}val-2010 task 8: Multi-way classification of semantic
  relations between pairs of nominals,''
\newblock in {\em Proc. of SemEval}, 2010.

\bibitem{li-etal-2023-well}
Jiang Li, Xiangdong Su, Xinlan Ma, and Guanglai Gao,
\newblock ``How well apply simple {MLP} to incomplete utterance rewriting?,''
\newblock in {\em Proc. of ACL}, 2023.

\bibitem{anonymous2023rethinking}
Li~Zhengdao, Cao Yong, Shuai Kefan, Miao Yiming, and Hwang Kai,
\newblock ``Rethinking the effectiveness of graph classification datasets in
  benchmarks for assessing {GNN}s,''
\newblock in {\em Submitted to The Twelfth ICLR}, 2023.

\bibitem{wang2022revisiting}
Yizhou Wang, Shixiang Tang, Feng Zhu, Lei Bai, Rui Zhao, Donglian Qi, and Wanli
  Ouyang,
\newblock ``Revisiting the transferability of supervised pretraining: an mlp
  perspective,''
\newblock in {\em Proc. of CVPR}, 2022.

\bibitem{vaswani2017attention}
Ashish Vaswani, Noam Shazeer, Niki Parmar, Jakob Uszkoreit, Llion Jones,
  Aidan~N Gomez, {\L}ukasz Kaiser, and Illia Polosukhin,
\newblock ``Attention is all you need,''
\newblock {\em Proc. of NeurIPS}, 2017.

\bibitem{devlin-etal-2019-bert}
Jacob Devlin, Ming-Wei Chang, Kenton Lee, and Kristina Toutanova,
\newblock ``{BERT}: Pre-training of deep bidirectional transformers for
  language understanding,''
\newblock in {\em Proc. of NAACL}, 2019.

\bibitem{ravishankar-etal-2021-attention}
Vinit Ravishankar, Artur Kulmizev, Mostafa Abdou, Anders S{\o}gaard, and Joakim
  Nivre,
\newblock ``Attention can reflect syntactic structure (if you let it),''
\newblock in {\em Proc. of EACL}, 2021.

\bibitem{liu-etal-2019-linguistic}
Nelson~F. Liu, Matt Gardner, Yonatan Belinkov, Matthew~E. Peters, and Noah~A.
  Smith,
\newblock ``Linguistic knowledge and transferability of contextual
  representations,''
\newblock in {\em Proc. of NAACL}, 2019.

\bibitem{niu-etal-2022-using}
Jingcheng Niu, Wenjie Lu, Eric Corlett, and Gerald Penn,
\newblock ``Using roark-hollingshead distance to probe {BERT}{'}s syntactic
  competence,''
\newblock in {\em Proceedings of the Fifth BlackboxNLP Workshop on Analyzing
  and Interpreting Neural Networks for NLP}, 2022.

\bibitem{conneau-etal-2018-cram}
Alexis Conneau, German Kruszewski, Guillaume Lample, Lo{\"\i}c Barrault, and
  Marco Baroni,
\newblock ``What you can cram into a single {\$}{\&}!{\#}* vector: Probing
  sentence embeddings for linguistic properties,''
\newblock in {\em Proc. of ACL}, 2018.

\bibitem{conneau-kiela-2018-senteval}
Alexis Conneau and Douwe Kiela,
\newblock ``{S}ent{E}val: An evaluation toolkit for universal sentence
  representations,''
\newblock in {\em Proc. of LREC}, 2018.

\end{thebibliography}
